\pdfoutput=1

\documentclass[11pt]{article}

\usepackage[preprint]{acl}

\usepackage{times}
\usepackage{latexsym}

\usepackage[T1]{fontenc}

\usepackage[utf8]{inputenc}

\usepackage{microtype}

\usepackage{inconsolata}

\usepackage[pdftex]{graphicx}

\usepackage{amsmath}
\usepackage{multirow}
\usepackage{makecell}
\usepackage{bm}
\usepackage{wrapfig}
\usepackage{enumitem}
\usepackage{tabularray}
\usepackage{booktabs}
\usepackage{amssymb}
\usepackage{pifont}
\usepackage{fontawesome5}
\usepackage{marvosym}

\usepackage{xcolor}
\usepackage{soul}

\newcommand{\modelname}{PaMis}

\title{Listening to Patients: A Framework of Detecting and Mitigating \\Patient Misreport for Medical Dialogue Generation
}

\author{%
  Lang Qin \\
  \texttt{qinlang14@mail.nankai.edu.cn} \\
  \And
  Yao Zhang \\
  \texttt{yaozhang@nankai.edu.cn} \\
  \AND
  Hongru Liang \\
  \texttt{lianghongru@scu.edu.cn} \\
  \And
  Adam Jatowt \\
  \texttt{Adam.Jatowt@uibk.ac.at} \\
  \And
  Zhenglu Yang \\
  \texttt{yangzl@nankai.edu.cn} \\
}

\begin{document}

\maketitle

\begin{abstract}
    \label{abstract}

    Medical Dialogue Systems aim to provide automated healthcare support through patient-agent conversations.
    Previous efforts typically regard patients as ideal users---one who accurately and consistently reports their health conditions. 
    However, in reality, patients often misreport their symptoms, leading to discrepancies between their reports and actual health conditions.
    Overlooking patient misreport will affect the quality of healthcare consultations provided by MDS.
    To address this issue, we argue that MDS should ``listen to patients'' and tackle two key challenges: how to detect and mitigate patient misreport effectively.
    In this work, we propose~\textbf{\modelname{}}, a framework of detecting and mitigating \underline{Pa}tient \underline{Mis}report for medical dialogue generation.
    \modelname{} first constructs dialogue entity graphs, then detects patient misreport based on graph entropy, and mitigates patient misreport by formulating clarifying questions.
    Experiments indicate that \modelname{} effectively enhances medical response generation, enabling models like GPT-4 to detect and mitigate patient misreports, and provide high-quality healthcare assistance.

    \end{abstract}

\section{Introduction}
\label{introduction}

\begin{figure}[t]
    \centering
    \includegraphics[width=1.0\linewidth]{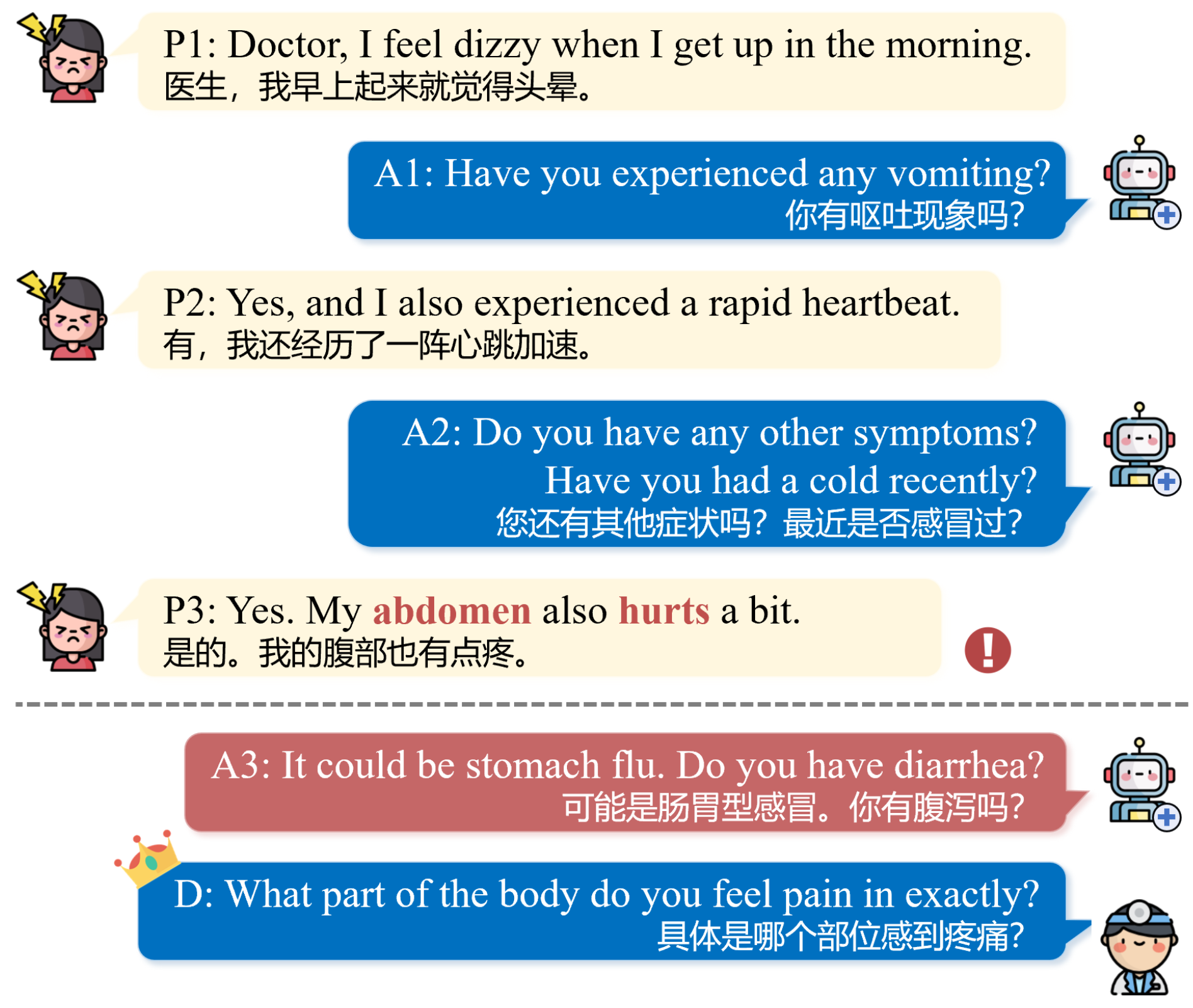}
    \caption{Example of patient misreport in patient (P)-agent (A) conversations and a response generated by the real experienced doctor (D).
    When the patient with myocardial infarction misreports angina as abdominal pain, the doctor remains vigilant and asks more detailed questions to discern the patient's actual symptoms.
  However, the agent can easily be influenced by the patient's misreport and arbitrarily shift the focus to stomach flu.}
    \label{fig:intro_example} 
\end{figure}

Medical Dialogue Systems (MDSs) aim to provide automated healthcare support through natural language interactions between patients and agents~\cite{li2021semi, liu2022nose,xu-etal-2024-reasoning}. 
The patient describes symptoms or health concerns, while the agent processes the patient self-report and responds with appropriate medical guidance and follow-up questions, mimicking the strategies employed by real doctors.
Taking Figure~\ref{fig:intro_example} as an example, when a patient reports that he/she \textit{feels dizzy (P1)}, the agent will inquire about possessing more related symptoms, such as \textit{vomiting (A1)} and \textit{cold (A2)}.
In order to better provide medical support,
previous works on MDSs has devoted significant effort to leveraging advanced frameworks (e.g., reinforcement learning~\cite{wei-etal-2018-task} and graph-based structures~\cite{lin-etal-2019-enhancing}) and integrating external medical knowledge~\cite{xu-etal-2023-medical}.

Despite extensive research, current efforts still operate under the assumption of an ideal patient---one who accurately and consistently reports their health conditions. 
In reality, patients often misreport their symptoms due to cognitive limitations or emotional factors, etc., leading to discrepancies between their reports and actual health conditions~\citep{berkman2011low, prior2011talking}.
Some research indicates that patients misreport their symptoms in approximately 15–20\% of cases~\cite{fleischer2015inattentive,merckelbach2019patients}.
Meanwhile, we observed that patient misreports occurred in 16.9\% of the dialogues within the public corpus~\citep{liu2022meddg}.
Still taking Figure~\ref{fig:intro_example} as an example, a patient experiencing a myocardial infarction might inaccurately describe angina as \textit{abdomen pain (P3)}. 
When by default the patient is able to accurately report his or her symptoms, the agent will arbitrarily infer that the patient may have a \textit{stomach flu (A3)}.
This would potentially delay appropriate treatment and adversely impact the patient's health.
In contrast, an experienced doctor would remain vigilant and ask \textit{more details about the location of pain (D)} to discern the patient's actual symptoms.

Therefore, it is essential to move beyond the assumption of idealized patients, enabling dialogue systems to detect potential misreports in patient narratives and ask clarifying questions, much like an experienced doctor, to discern the actual health conditions.
When confronting the issue of patient misreport, we propose that an effective MDS should tackle two key challenges:

\begin{itemize}[leftmargin=*] 
  \setlength{\itemsep}{-2pt}
       \item \textit{\textbf{Misreport Detection}}: Due to the complexity and subtlety of patient narratives, misreport often emerges as implicit contradictions in patients' self-reported information. Detecting patient misreports may span multiple dialogue rounds and vary across individuals, as well as requiring deep medical domain knowledge.

       \item \textit{\textbf{Misreport Mitigation}}:
       If detecting a misreport, the agent needs to generate targeted clarifying questions to help the patient calibrate the self-report to ensure high-quality healthcare. The generated questions need to integrate the detected misreport with medical knowledge to not only mitigate misreporting, but also maintain natural dialogue flow.
  \end{itemize}

  Inspired by the aforementioned analyses, we propose~\textbf{\modelname{}}, a framework of detecting and mitigating \underline{Pa}tient \underline{Mis}report for medical dialogue generation.
  \modelname{} utilizes dialogue context and external medical knowledge to (1) construct entity graph, (2) detect patient misreport based on dialogue entity graph and (3) mitigate patient misreport by formulating clarifying questions.
  Specifically, the misreport detection module calculate the structural entropy of the dialogue entity graph for detection, while the misreport mitigation module generates controlled clarifying questions based on the detected misreport information.
  Experimental results on two medical dialogue datasets, MedDG and KaMed, demonstrate \modelname{}'s superior performance in medical response generation. 
  Furthermore, when integrated with state-of-the-art language models like GPT-4, \modelname{} significantly enhances their ability to detect and mitigate patient misreports.
On the strength of the encouraging performance, we are confident that \modelname{} can effectively contribute to the MDS in providing high-quality healthcare assistance.
In conclusion, the key contributions of this research are outlined below.


  \begin{itemize}[leftmargin=*] 
    \setlength{\itemsep}{-2pt}
         \item We call attention to the underexplored phenomenon of patient misreport that occurs in patient-agent conversations.
         \item We propose~\modelname{}, a framework for detecting patient misreport based on graph entropy, and mitigating patient misreport by formulating clarifying questions.
         \item Experiments indicate that \modelname{} effectively enhances medical response generation, enabling models like GPT-4 to detect and mitigate patient misreports, and provide high-quality healthcare assistance.
    \end{itemize}

\section{Related Work}
\label{relatedwork}
\begin{figure*}[t]
\setlength{\belowcaptionskip}{-15pt} 
  \centering
  \includegraphics[width=1.0\textwidth]{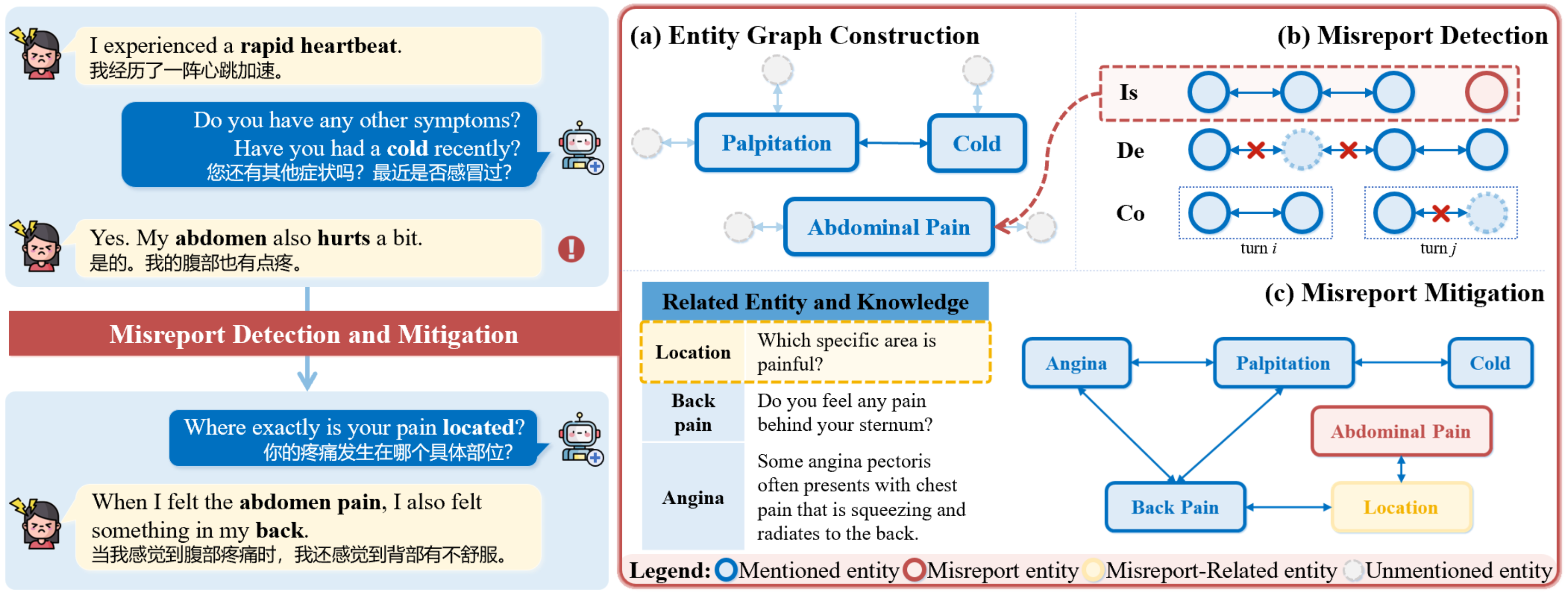}
    \caption{
    An illustration of~\modelname, using the dialogue in Figure~\ref{fig:intro_example} as an example. \modelname{} first constructs the entity graph, and then detects and mitigates the patient misreport based on the entity graph.
 }
  \label{fig:model} 
\end{figure*}

\subsection{Medical Dialogue Systems}
Medical dialogue systems aim to collect symptoms and automate diagnosis by obtaining information about patients' health conditions through conversation.
\noindent
\textbf{(1) Symptom Collection:}
Given the critical role of entities in medical dialogues~\citep{liu2022meddg}, previous studies have developed entity-aware models for symptom collection.
\citet{lin-etal-2019-enhancing} utilized a symptom graph with global attention to identify symptoms. \citet{xu-etal-2023-medical} introduced a framework that combines medical entity flows with dialogue action flows.
\noindent
\textbf{(2) Automatic Diagnosis:}
Early research focused on developing agents that mimic physicians' diagnostic dialogue behavior. \citet{wei-etal-2018-task} applied DQN to refine strategies for extracting symptoms from dialogues, aiding in diagnosis. \citet{xu2019end} incorporated knowledge graphs to optimize end-to-end automated diagnosis.
\noindent
\textbf{(3) MDS Challenges:}
Previous studies have highlighted the effectiveness of guiding agents to emulate physician behavior~\citep{li2021semi, liu2022nose} and have addressed challenges such as vague patient statements and limited medical data~\citep{zhao2022medical, xu-etal-2023-medical, tang2023term, lin2021graph, hou2023mtdiag}.
However, they have not adequately considered the issue of patient misreporting, and recent research continues to assume that patients will provide accurate answers based on correct facts~\citep{li2024mediqquestionaskingllmsbenchmark}.

This study addresses patient misreporting by introducing a medical dialogue management framework that detects and mitigates inaccuracies.

\subsection{Misreport in Task-oriented Dialogue}
To address vague or incorrect information provided by users, task-oriented dialogue systems often employ proactive questioning to clarify issues.
These systems construct clarifying questions to maximize the expected information gain based on the principle that ``a good question is one whose expected answer is the most useful~\citep{rao2017you}.''
Specifically, when confronted with unclear user intent, ambiguous expressions, or statements containing errors, prior research suggests several approaches: 
\citet{Aliannejadi2019asking} retrieve related questions based on interaction history to identify user information needs in open-domain information-seeking tasks;
\citet{xu-etal-2019-asking} employ a ``judge-then-generate'' method in knowledge-based QA systems to resolve ambiguities in knowledge items;
\citet{feng2023towards} utilize clarifying questions to gather necessary user-specific information in task-oriented dialogues;
\citet{zhao2024generating} address ambiguous or multifaceted user intents in web search through multi-turn questions.
Additionally, \citet{Oshima_2023_ICCV} investigated challenges arising from human errors that lead to agent failures in goal-oriented visual QA tasks, emphasizing the importance of agents detecting and pointing out these errors.

In this paper, we propose a misreport detection mechanism for medical dialogue systems and utilize agent responses to disambiguate information and accurately capture the patient’s health status.

\section{Methodology}
\label{method}

\subsection{Overview}
\label{sec:overview}

We formulate the doctor-patient conversation as $\mathcal{C}=\left\{\left(\mathcal{P}_{i}, \mathcal{D}_{i}\right)\right\}_{i=1}^{T}$, where $\mathcal{P}_{i}$ denotes the patient's statement and $\mathcal{D}_{i}$ represents the doctor's response.
The primary objective of MDS is to develop a physician agent that generates an appropriate response $\mathcal{D}_{i}$ based on the dialogue history $\mathcal{C}_{1:i-1}$ and the patient's current statement $\mathcal{P}_{i}$.

The entities $\{{e}_{i}\}$ appearing in the conversation $\mathcal{C}$ can be organized into a graph ${G}_{e}$ to represent the patient’s health condition and the doctor’s logic of inquiry.
In addition to generating responses $\mathcal{D}_{i}$, a MDS that addresses patient misreports must detect misreported entities ${e}_{m}$ and ask clarifying questions to reduce inaccuracies.
The system leverages both the dialogue content and the entity graph to maximize the probability $P\left(\mathcal{D}_{i} \mid \mathcal{C}_{1: i-1}, \mathcal{P}_{i}, {G}_{e}, {e}_{m}\right)$.

\subsection{Entity Graph Construction}
\label{sec:graph}
We introduce a modeling approach that integrates a static knowledge graph $G$ with a dynamic dialogue entity graph ${G}_{e} \subseteq G$ to improve detection effectiveness.
The graph $G$ is constructed from the corpus, with nodes representing entities and edge weights representing co-occurrence relationships.
The likelihood of mentioning entity $e_{t}$ at turn $t$ is modeled as the conditional probability given the entities $\left\{e_{1}, e_{2}, \ldots, e_{t-1}\right\}$ in the dialogue history:
\begin{equation}
    P\left(e_{t} \mid \mathcal{C}_{1: t-1}\right) \propto P\left(e_{t} \mid e_{1}, e_{2}, \ldots, e_{t-1}\right).
\end{equation}
Thus, we use the co-occurrence frequency in the corpus as the directed\footnote{Based on the sequence of entity occurrences in the corpus, it reflects the logic of doctors' inquiries.} weight $w_{ij}$ from $e_{i}$ to $e_{j}$:
\begin{equation}
    w_{i j}=\frac{\operatorname{freq}\left(e_{i}, e_{j}\right)}{\operatorname{freq}\left(e_{i}\right)} \propto P\left(e_{j} \mid e_{i}\right).
\end{equation}
The edge weights in graph $G$ serve as the basis for ranking relevant entities during subsequent retrieval for response generation.

The dialogue history references or denies medical entities, which are extracted through medical slot-filling~\citep{hu2023matching}. 
The extracted entities and relationships $r=(e_{i}, e_{j}, w_{i j})$ must align with established background knowledge $G$, rather than transient associations.
Entities identified in each utterance form a separate dialogue entity graph ${G}_{e}$ for that utterance.
We track changes in these graphs to generate a sequence of dynamic graphs, $\left\{G_{e_{1}}, G_{e_{2}}, \ldots\right\}$.
Misreport detection and classification are then performed using graph entropy theory.\\

\subsection{Misreport Detection}
\label{sec:detection}
Patient misreporting occurs when discrepancies arise between the health conditions described in the dialogue and the patient’s actual health conditions.
Health conditions often follow inherent medical co-occurrence relationships~\citep{NEURIPS2022_RE3, NEURIPS2023_REFINE}, which can be disrupted by misreports, leading to anomalies in the dialogue’s entity graph.
We analyze misreport patterns using real-world data~\citep{liu2022meddg, li2021semi} and apply graph entropy to detect these misreports.\\

\noindent
\textbf{(1) Misreport Feature Classification}

Patient misreports often lack clear linguistic indicators, but they manifest as anomalies on the entity graph due to inconsistencies with established medical knowledge.
Based on observed patient behavior, we categorize misreports into three types.

\begin{itemize}[leftmargin=*]
\setlength{\itemsep}{-2pt}
    \item \textit{Introducing isolated entities} (Figure~\ref{fig:model}(b)-$(Is)$).
    This type is typically initiated by patients because of anxiety or vague descriptions.
    

    \item \textit{Denying crucial entities} (Figure~\ref{fig:model}(b)-$(De)$).
    This type involves deleting entities and transferring graphs to disconnect components, which occurs when patients deny entities in doctors' inquiries.
    \item \textit{Presenting self-contradictions} (Figure~\ref{fig:model}(b)-$(Co)$).
    This type involves entities appearing and disappearing without disrupting the graph's structure, typically reflecting patients misreporting the duration of symptoms or medical history\footnote{Doctors may focus on recent symptoms for acute conditions or long-term medication history for chronic illnesses. Patients might provide contradictory information, such as first claiming ``never had a stomach problem'' and later mentioning ``occasional stomach pain.''}.
\end{itemize}

\noindent
\textbf{(2) Graph Entropy-based Detection}

Graph entropy characterizes the structural information of graphs.
Given that misreporting disrupts graph structure, we use graph entropy for detection.
Specifically, to address the potential disconnected features of entity graphs, we apply one-dimensional structural entropy \citep{li2016structural} to detect misreporting phenomena.

The definition of graph entropy aligns with Shannon's information entropy \citep{Shannon}, aiming to represent structure information and complexity through the degree distribution of nodes.
In the following formula, ${d}_{i}$ denotes the degree of node $i$, and $\operatorname{Vol}(G)$ is the degree sum of $G$:
\begin{equation}
    H(G)=-\sum_{i=1}^{n} \frac{d_{i}}{\operatorname{Vol}(G)} \log _{2} \frac{d_{i}}{\operatorname{Vol}(G)}.
\end{equation}

For the entity graph $G_e$, the one-dimensional structural entropy is calculated as the weighted average of the entropy of each connected component, defined as follows: 
\begin{equation}
H^{1}(G)=\frac{1}{\operatorname{Vol}(G)} \sum_{j=1}^{L} \operatorname{Vol}\left(G_{j}\right) \cdot H^{1}\left(G_{j}\right),
\end{equation}
where $G_j$ represents a connected subgraph of $G$, $L$ represents the number of connected components, and the entropy is considered as $0$ if has no edges.

An effective inquiry process should enhance the co-occurrence relationships between entities, thereby expanding information pathways and increasing graph entropy.
As illustrated in Figure~\ref{fig:model}, misreports can disrupt this pattern.
A single calculation of graph entropy can be used to detect and classify such misreports:

\begin{itemize}[leftmargin=*]
\setlength{\itemsep}{-2pt}
    \item \textit{(Is)} occurs when the number of nodes increases while the graph entropy remains unchanged because an isolated node has zero entropy.

    \item \textit{(De)} and \textit{(Co)} reduce information pathways in the graph, leading to a decrease in entropy.
    When nodes are lost, the entropy lower bound for contradictions is higher than the upper bound for denials.
    This boundary helps detect and classify misreports, with the formulations provided below and detailed proofs available in Appendix~\ref{appendix_entropy}:
    
\begin{equation}
\label{eq_low}
    \small H(G_{\textit{Co}})\geq-\sum_{i=1}^{n} \frac{d_{i}-1}{2(n-1)} \log _{2} \frac{d_{i}-1}{2(n-1)},
\end{equation}

\begin{equation}
\label{eq_upper}
    \small H(G_{\textit{De}})\leq-\left( \sum_{i=1}^{n-2} F(d_i) + F(d_{n-1} - 1) \right),
\end{equation}

\begin{equation}
\label{eq_upper_2}
     \small F(d) =\frac{d}{\operatorname{Vol}(G) - 4} \log_{2} \frac{d}{\operatorname{Vol}(G) - 4}.
\end{equation}

\end{itemize}

In summary, the sequence of graph entropy values enables the straightforward detection and classification of misreports.

\subsection{Misreport Mitigation}
\label{sec:mitigation}

The mitigation module generates clarifying questions based on detected patient misreports to obtain accurate information about patients' health conditions and mitigate misreporting.\\

\noindent
\textbf{(1) Response Generation}

Misreport mitigation can be perceived as an instance of response generation as it involves the agent asking questions.
Therefore, we first introduce the standard process of~\modelname~in utilizing medical knowledge to generate responses.
As illustrated in Figure~\ref{fig:model}, we establish a connection between the medical knowledge graph and real physician responses based on entities.
Next, we identify the most valuable responses to serve as guiding information.
A set of real responses $\{S\}$ containing the entity $e$ can be extracted from the corpus.
We calculate the cosine similarity of word vectors $\mathbf{v}$ for the sentences in $\{S\}$ and select the top-k sentences that exhibit the highest average similarity to others as the knowledge $\{S_e\}$related to entity $e$:
\begin{equation}
    \small \operatorname{avg\_ sim}(e)=\frac{1}{|S|} \sum_{s_{i}, s_{j} \in S, i \neq j} \operatorname{sim}\left(s_{i}, s_{j}\right),
\end{equation}

\begin{equation}
    \small \operatorname{sim}\left(s_{i}, s_{j}\right)=\frac{\mathbf{v}_{i} \cdot \mathbf{v}_{j}}{\left\|\mathbf{v}_{i}\right\| \mathbf{v}_{j} \|}.
\end{equation}
The agent leverages the dialogue context and the dialogue entity graph to retrieve knowledge pairs $\{(e, S_e)\}$ from the one-hop neighbors of existing nodes.
The candidate ranking process uses the weights described in Section~\ref{sec:graph}, where the edge weights to adjacent nodes serve as scores for neighboring entities.
This step does not mean performing the entity prediction task in MDS but provides a preliminary enhancement to the proposed framework.
This explicit approach gives the framework flexibility to incorporate generative models, including large language models.

Overall, the aim of response generation is to maximize the conditional probability $P\left(\mathcal{D}_{i} \mid \mathcal{C}_{1: i-1}, \mathcal{P}_{i}, \{(e, S_e)\}\right)$.\\

\noindent
\textbf{(2) Clarifying Process}

Based on the approach described above, the clarifying process is triggered when misreports are detected.
This process involves re-retrieving entity-response pairs $\{(e, S_e)\}$ related to the misreported entity $e_m$ as guiding information.
The implications vary depending on the scenario:
\begin{itemize}[leftmargin=*] 
    \setlength{\itemsep}{-2pt}
    \item $(Is)$
    \modelname~retrieves bridging nodes from the static graph $G$ to complete the dialogue entity graph, aiming to generate the next inquiry that restores the graph to a connected state. 
    \item $(De)$
    \modelname~uses related knowledge of denied entities to generate clarifying questions, reconfirming the existence or absence of the relevant health condition.
    \item $(Co)$
    \modelname~selects broader attribute entities (such as duration, location, etc.) to generate inquiries, obtaining supplementary information to assist in judgment.
\end{itemize}

In summary, related knowledge is integrated to augment clarifying question generations by retrieving information related to misreports.

\begin{table}[t]
    \centering
    \resizebox{0.40\textwidth}{!}{
    \begin{tabular}{l|cc}
        \toprule[1.5pt]
        \textbf{Dataset} & MedDG & KaMed
        \\
        \midrule[1.2pt]
        \textbf{ Dialogues } & 17,864 & 63,754 \\
        \text{Avg. \# of utterances} &19.85 & 23.25
        \\
        \text{Avg. \# of entities} &12.11 & 14.94
        \\
        \bottomrule[1.5pt]
    \end{tabular}}
    \caption{The statistics of datasets.}
    \label{tab:statistics}
\end{table}

\section{Experiments}
\label{experimetns}

We conduct experiments focusing on three core research questions:
\begin{itemize}[leftmargin=*]
\setlength{\itemsep}{-2pt}
    \item \textbf{RQ1}: Does \modelname{} outperform current methods in terms of overall performance of medical response generation?
    \item \textbf{RQ2}: Does \modelname{} perform better in meeting the fundamental requirements of the medical dialogue system?
    \item \textbf{RQ3}: Can \modelname{} effectively reduce misreports in doctor-patient interactions?
\end{itemize}

\begin{table*}[t]
\setlength{\belowcaptionskip}{-15pt} 
    \centering
    \setlength\tabcolsep{3pt}
    \resizebox{0.99\textwidth}{!}{
    \begin{tabular}{l|cccccc|cccccc}
        \toprule[1.6pt]
        \specialrule{0em}{1pt}{1pt}
        \multicolumn{1}{l|}{\multirow{2}{*}{\textbf{Method}}} &
        \multicolumn{6}{c|}{\textbf{MedDG}}                  &
        \multicolumn{6}{c}{\textbf{KaMed}}                
        \\
        \cmidrule(lr){2-7}\cmidrule(lr){8-13}
        \multicolumn{1}{l|}{}  &
        \multicolumn{1}{c}{\textbf{P}} &
        \multicolumn{1}{c}{\textbf{A}} &
        \multicolumn{1}{c}{\textbf{H}} &
        \multicolumn{1}{c}{\textbf{LQ}} &
        \multicolumn{1}{c}{\textbf{DV}} &
        \multicolumn{1}{c|}{\textbf{Overall}}  &
        \multicolumn{1}{c}{\textbf{P}} &
        \multicolumn{1}{c}{\textbf{A}} &
        \multicolumn{1}{c}{\textbf{H}} &
        \multicolumn{1}{c}{\textbf{LQ}} &
        \multicolumn{1}{c}{\textbf{DV}} &
        \multicolumn{1}{c}{\textbf{Overall}}

        \\
        \specialrule{0em}{1pt}{1pt}
        \midrule[1.0pt]
        \specialrule{0em}{1pt}{1pt}
HuatuoGPT2	
&\makecell[c]{3.46\\[-0.5pt]\footnotesize{±0.03}} &\makecell[c]{3.99\\[-0.5pt]\footnotesize{±0.02}} &\makecell[c]{3.32\\[-0.5pt]\footnotesize{±0.03}} &\makecell[c]{4.27\\[-0.5pt]\footnotesize{±0.07}} &\makecell[c]{3.08\\[-0.5pt]\footnotesize{±0.03}} &\makecell[c]{18.11\\[-0.5pt]\footnotesize{±0.08}}

&\makecell[c]{3.35\\[-0.5pt]\footnotesize{±0.05}} &\makecell[c]{3.95\\[-0.5pt]\footnotesize{±0.05}} &\makecell[c]{3.26\\[-0.5pt]\footnotesize{±0.04}} &\makecell[c]{4.18\\[-0.5pt]\footnotesize{±0.03}} &\makecell[c]{2.96\\[-0.5pt]\footnotesize{±0.04}} &\makecell[c]{17.70\\[-0.5pt]\footnotesize{±0.19}}
\\
DISC-MedLLM
&\makecell[c]{3.20\\[-0.5pt]\footnotesize{±0.05}} &\makecell[c]{3.95\\[-0.5pt]\footnotesize{±0.05}} &\makecell[c]{3.19\\[-0.5pt]\footnotesize{±0.06}} &\makecell[c]{4.16\\[-0.5pt]\footnotesize{±0.09}} &\makecell[c]{2.87\\[-0.5pt]\footnotesize{±0.07}} &\makecell[c]{17.37\\[-0.5pt]\footnotesize{±0.28}}

&\makecell[c]{3.12\\[-0.5pt]\footnotesize{±0.01}} &\makecell[c]{3.93\\[-0.5pt]\footnotesize{±0.02}} &\makecell[c]{3.10\\[-0.5pt]\footnotesize{±0.01}} &\makecell[c]{4.03\\[-0.5pt]\footnotesize{±0.05}} &\makecell[c]{2.75\\[-0.5pt]\footnotesize{±0.04}} &\makecell[c]{16.93\\[-0.5pt]\footnotesize{±0.08}}
\\ \midrule[0.5pt]
ChatGPT
&\makecell[c]{3.70\\[-0.5pt]\footnotesize{±0.05}} &\makecell[c]{4.15\\[-0.5pt]\footnotesize{±0.03}} &\makecell[c]{3.61\\[-0.5pt]\footnotesize{±0.05}} &\makecell[c]{4.23\\[-0.5pt]\footnotesize{±0.01}} &\makecell[c]{3.43\\[-0.5pt]\footnotesize{±0.04}} &\makecell[c]{19.13\\[-0.5pt]\footnotesize{±0.17}}

&\makecell[c]{3.72\\[-0.5pt]\footnotesize{±0.03}} &\makecell[c]{4.18\\[-0.5pt]\footnotesize{±0.04}} &\makecell[c]{3.61\\[-0.5pt]\footnotesize{±0.05}} &\makecell[c]{4.19\\[-0.5pt]\footnotesize{±0.00}} &\makecell[c]{3.44\\[-0.5pt]\footnotesize{±0.04}} &\makecell[c]{19.14\\[-0.5pt]\footnotesize{±0.14}}
\\
ChatGPT \footnotesize{(w/ Gold Know.)}
&\makecell[c]{3.90\\[-0.5pt]\footnotesize{±0.02}} &\makecell[c]{4.23\\[-0.5pt]\footnotesize{±0.02}} &\makecell[c]{3.75\\[-0.5pt]\footnotesize{±0.06}} &\makecell[c]{4.28\\[-0.5pt]\footnotesize{±0.02}} &\makecell[c]{3.70\\[-0.5pt]\footnotesize{±0.04}} &\makecell[c]{19.87\\[-0.5pt]\footnotesize{±0.12}}

&\makecell[c]{3.89\\[-0.5pt]\footnotesize{±0.06}} &\makecell[c]{4.23\\[-0.5pt]\footnotesize{±0.07}} &\makecell[c]{3.77\\[-0.5pt]\footnotesize{±0.06}} &\makecell[c]{4.24\\[-0.5pt]\footnotesize{±0.03}} &\makecell[c]{3.65\\[-0.5pt]\footnotesize{±0.05}} &\makecell[c]{19.78\\[-0.5pt]\footnotesize{±0.27}}
\\\midrule[0.5pt]
GPT-4
&\makecell[c]{3.93\\[-0.5pt]\footnotesize{±0.03}} &\makecell[c]{4.21\\[-0.5pt]\footnotesize{±0.01}} &\makecell[c]{3.82\\[-0.5pt]\footnotesize{±0.01}} &\makecell[c]{4.36\\[-0.5pt]\footnotesize{±0.03}} &\makecell[c]{3.61\\[-0.5pt]\footnotesize{±0.02}} &\makecell[c]{19.93\\[-0.5pt]\footnotesize{±0.04}}

&\makecell[c]{3.98\\[-0.5pt]\footnotesize{±0.03}} &\makecell[c]{4.27\\[-0.5pt]\footnotesize{±0.03}} &\makecell[c]{3.89\\[-0.5pt]\footnotesize{±0.05}} &\makecell[c]{4.37\\[-0.5pt]\footnotesize{±0.02}} &\makecell[c]{3.65\\[-0.5pt]\footnotesize{±0.04}} &\makecell[c]{20.16\\[-0.5pt]\footnotesize{±0.14}}
\\
GPT-4 \footnotesize{(w/ Gold Know.)}
&\makecell[c]{3.96\\[-0.5pt]\footnotesize{±0.02}} &\makecell[c]{4.23\\[-0.5pt]\footnotesize{±0.02}} &\makecell[c]{3.87\\[-0.5pt]\footnotesize{±0.03}} &\makecell[c]{\textbf{4.37}\\[-0.5pt]\footnotesize{±0.02}} &\makecell[c]{3.67\\[-0.5pt]\footnotesize{±0.03}} &\makecell[c]{20.10\\[-0.5pt]\footnotesize{±0.09}}

&\makecell[c]{4.03\\[-0.5pt]\footnotesize{±0.04}} &\makecell[c]{4.29\\[-0.5pt]\footnotesize{±0.02}} &\makecell[c]{3.96\\[-0.5pt]\footnotesize{±0.04}} &\makecell[c]{\textbf{4.38}\\[-0.5pt]\footnotesize{±0.02}} &\makecell[c]{3.72\\[-0.5pt]\footnotesize{±0.05}} &\makecell[c]{20.38\\[-0.5pt]\footnotesize{±0.14}}
\\
\specialrule{0em}{1pt}{1pt}
\midrule[0.8pt]
\specialrule{0em}{1pt}{1pt}

\modelname{}
&\makecell[c]{\textbf{4.00}$^\dagger$\\[-0.5pt]\footnotesize{±0.01}} &\makecell[c]{\textbf{4.26}$^\dagger$\\[-0.5pt]\footnotesize{±0.02}} &\makecell[c]{\textbf{3.93}$^\dagger$\\[-0.5pt]\footnotesize{±0.05}} &\makecell[c]{4.36\\[-0.5pt]\footnotesize{±0.01}} &\makecell[c]{\textbf{3.78}$^\dagger$\\[-0.5pt]\footnotesize{±0.02}} &\makecell[c]{\textbf{20.33}$^\dagger$\\[-0.5pt]\footnotesize{±0.03}}

&\makecell[c]{\textbf{4.05}\\[-0.5pt]\footnotesize{±0.02}} &\makecell[c]{\textbf{4.32}\\[-0.5pt]\footnotesize{±0.01}} &\makecell[c]{\textbf{4.00}\\[-0.5pt]\footnotesize{±0.02}} &\makecell[c]{4.37\\[-0.5pt]\footnotesize{±0.03}} &\makecell[c]{\textbf{3.84}$^\dagger$\\[-0.5pt]\footnotesize{±0.01}} &\makecell[c]{\textbf{20.58}$^\dagger$\\[-0.5pt]\footnotesize{±0.06}}

        \\
        \specialrule{0em}{1pt}{1pt}
        \bottomrule[1.6pt]
    \end{tabular}}
    \caption{Evaluation results of responses generated from LLMs
    and \modelname{}.
    ``Gold'' indicates that the entities in the actual doctor's responses are included in the input.
    $\dagger$ denotes statistically significant differences (p < 0.05).}
    \label{tab:llms_responses}

\end{table*}
\begin{table}[t]
    \tiny
        \centering
        \setlength\tabcolsep{3pt}
        \resizebox{0.49\textwidth}{!}{
        \begin{tabular}{l|ccccccc}
            \toprule[1.2pt]

            \multicolumn{1}{l|}{\textbf{Method}}  &
            \multicolumn{1}{c}{\textbf{B-1}} &
            \multicolumn{1}{c}{\textbf{B-4}} &
            \multicolumn{1}{c}{\textbf{R-1}} &
            \multicolumn{1}{c}{\textbf{R-2}} &
            \multicolumn{1}{c}{\textbf{D-1}} &
            \multicolumn{1}{c}{\textbf{D-2}} &
            \multicolumn{1}{c}{\textbf{E-F}}  
            \\\midrule[0.8pt]
            \multicolumn{8}{c}{\textbf{MedDG}} 
            \\
            \midrule[0.4pt]
            \text{GPT-2}
            &35.27 &19.16 &28.74 &13.61 & 0.87 & 11.20 & 16.14 
            
            \\
            \text{VRBot}
            &29.69 &16.34 &24.69 &11.23 & - & - &12.78
           
            \\
            \text{DFMED}
            &42.56 &22.53 &29.31 &14.21 & 0.85 & 8.28 &22.66
          
            \\
            \text{EMULATION}
            &42.35 &22.76 &\textbf{30.91} &\textbf{15.17} &- &- &24.03
           
            \\
            \midrule[0.4pt]
            \textbf{\modelname{}}
            &\textbf{44.28} &\textbf{24.88} &28.12 &13.80 & \textbf{1.23} & \textbf{11.58} & \textbf{25.13}
           
            \\
            \midrule[0.4pt]
            \multicolumn{8}{c}{\textbf{KaMed}} 
            \\\midrule[0.8pt]
            \text{GPT-2}
        
            &33.76 &17.82 &26.80 &10.56 &- &- &17.26
            \\
            \text{VRBot}
   
            &30.04 &16.36 &18.71 &7.28 &\textbf{1.61} &9.30 &12.08
            \\
            \text{DFMED}
       
            &40.50 &20.92 &28.33 &11.73 &- &- &22.31
            \\
            \text{EMULATION}
          
            &39.87 &19.79 &\textbf{28.54} &\textbf{12.33} &- &- &24.27
            \\
            \midrule[0.4pt]
            \textbf{\modelname{}}
      
            &\textbf{41.02} &\textbf{21.30} &28.42 &12.01 &1.30 &\textbf{11.39} &\textbf{28.18}
            \\
            \bottomrule[1.2pt]
        \end{tabular}
        }
        \caption{Evaluation results of responses generated from fine-tuned baselines and \modelname{} under the N-gram-based metrics. Baseline results are reported from \citet{li2021semi} and \citet{xu-etal-2024-reasoning}.}
        \label{tab:results_generation}
    \end{table}

\subsection{Settings}
\label{sec:settings}

\noindent
\textbf{Datasets}\quad
We conducted experiments using the MedDG dataset~\citep{liu2022meddg} and the KaMed dataset~\citep{li2021semi}.
Medical entities mentioned in the dialogues were annotated by domain experts.
Detailed dataset statistics are provided in Table~\ref{tab:statistics}.
To enrich the medical background knowledge, we integrated multiple knowledge graphs from the datasets, focusing on the co-occurrence relationships between medical entities.

\noindent
\textbf{Baselines}\quad
We selected two categories of baselines: LLMs with demonstrated medical capabilities and fine-tuned models. Detailed information about the baselines is provided in Appendix~\ref{appendix_baselines}.

(1) \textit{LLMs}: DISC-MedLLM~\citep{bao2023discmedllmbridginggenerallarge}, HuatuoGPT2~\citep{chen2024huatuogptii}, GPT-3.5~\citep{ouyang2022traininglanguagemodelsfollow}, and GPT-4~\citep{openai2024gpt4technicalreport}.

(2) \textit{Fine-tuned models}: GPT-2~\citep{radford2019language}, VRBot~\citep{li2021semi}, DFMED~\citep{xu-etal-2023-medical}, and EMULATION~\citep{xu-etal-2024-reasoning}.



\noindent
\textbf{Metrics}\quad
\label{sec:metrics}
We employed three categories of evaluation metrics:

(1) \textit{LLM-based}:
We follow previous studies~\citep{bao2023discmedllmbridginggenerallarge, xu-etal-2024-reasoning} that employ LLMs (e.g., GPT-4) as evaluators to assess MDS on the dimensions of \textit{Proactivity, Accuracy, Helpfulness, }and\textit{ Linguistic Quality}.
Additionally, we introduced a dimension called \textit{Diagnostic Vigilance}\footnote{DV: The doctor skillfully identifies inconsistencies or errors in the patient's statement and asks clarifying questions to ensure an accurate diagnosis.} to measure the model's ability to detect and respond to misreporting phenomena.

(2) \textit{N-gram-based}:
We follow prior works~\citep{liu2022meddg, xu-etal-2023-medical} that utilize BLEU~\citep{papineni-etal-2002-bleu}, ROUGE~\citep{lin-2004-rouge}, Distinct~\citep{li-etal-2016-diversity}, and Entity-F1~\citep{liu2022meddg} for the response generation task.

(3) \textit{Misreport-aware metrics}:
To evaluate the effectiveness of misreport mitigation, we introduce two metrics: \textit{$\Delta GE$}, which quantifies the average change in graph entropy before and after interactions, and \textit{MR}, the mitigation rate of misreports.

\noindent
\textbf{Implementation Details}\quad
\label{sec:implementation}
Building on previous studies~\citep{chen2023llmempowered, bao2023discmedllmbridginggenerallarge}, we construct prompts for generating responses and evaluations in comparison with LLMs.
To compare with fine-tuned models, we employ a backbone model\footnote{\url{huggingface.co/fnlp/bart-base-chinese}} of similar scale to the baselines.
The code will be released upon acceptance of this paper.
Details of the prompts and additional implementation information are provided in Appendix~\ref{appendix_implement} and \ref{appendix_interact}.

\begin{figure}[t]
  \centering
  \includegraphics[width=0.45\textwidth]{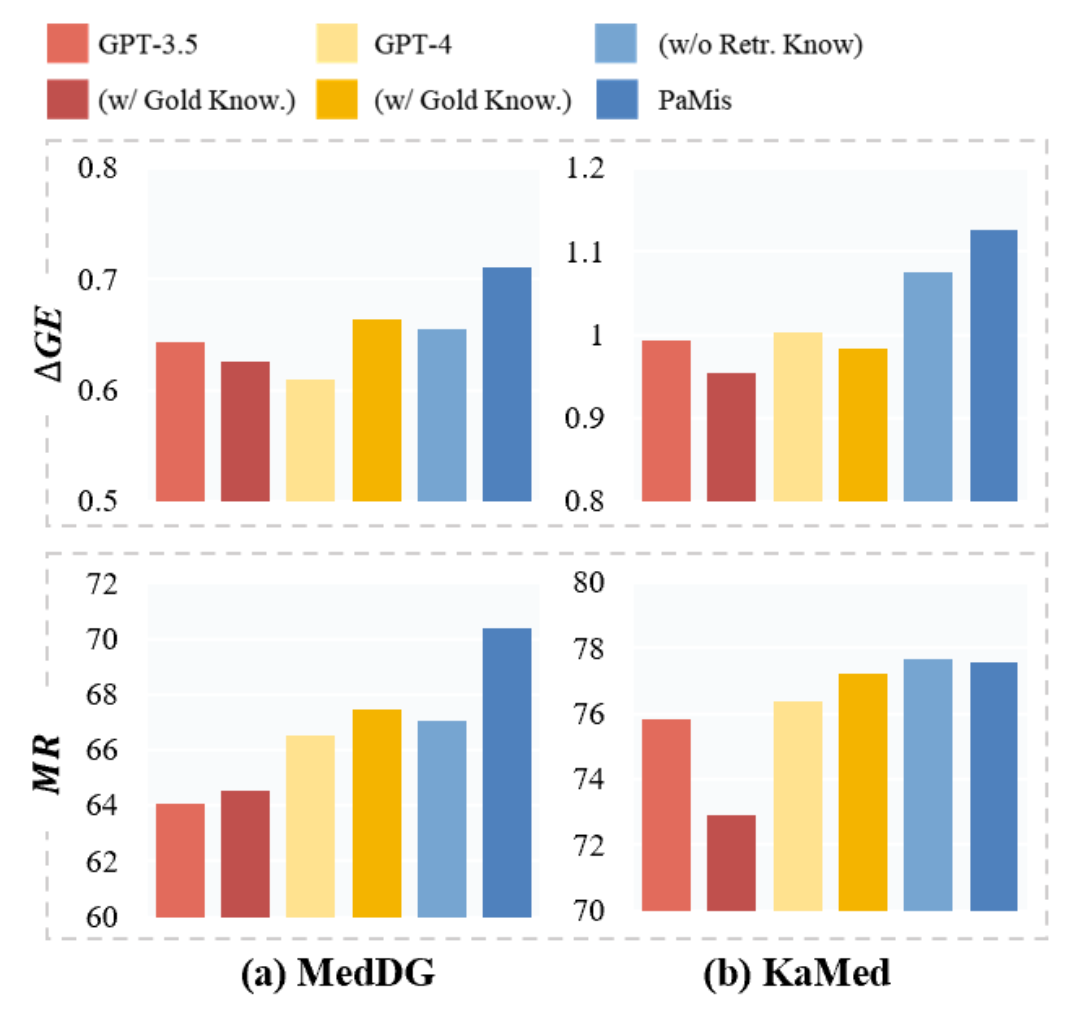}
    \caption{Evaluation results of interactive experiment under two misreport-aware metrics: \textit{$\Delta GE$} and \textit{MR}.
 }
  \label{fig:interact} 
\end{figure} 

\subsection{Results and Observations}
\label{sec:results}
\noindent
\textbf{Overall Performance}\quad
We conducted experiments on the MedDG and KaMed datasets with the response generation task, which involves the agent acting as a doctor and responding to the patient.
Results are presented in Tables~\ref{tab:llms_responses} and~\ref{tab:results_generation}.

For \textbf{RQ1}, Table~\ref{tab:llms_responses} demonstrates that while \modelname{} is specifically designed to address patient misreports, it outperforms advanced LLMs on general metrics.
We provided a robust baseline setting (i.e., w/ Gold Know.) for LLMs with medical capabilities.
Nevertheless, introducing the misreporting mechanism and related entities in the input content via \modelname{}~significantly improved response quality.
Given that GPT-4 served as the generation model, this suggests that the \modelname{}~management framework can enhance even highly advanced methods.
Furthermore, the improvements are concentrated in areas beyond linguistic quality, suggesting that the enhancements arise from medical-related capabilities rather than language tricks.

For \textbf{RQ2}, The n-gram-based results presented in Table~\ref{tab:results_generation} illustrate that the proposed misreport mitigation module can remain competitive with the state-of-the-art approach in the response generation task of medical dialogue systems.
Specifically, when relevant knowledge from the entity graph is introduced into the input of the generation model, it tends to generate inquiries that comprehensively incorporate related entities.
This approach improves the E-F score but reduces the ROUGE score compared to ground-truth responses, as some doctors prefer shorter, more conversational questions.

The aforementioned results indicate that \modelname{} is able to simulate real doctors who possess professional skills to manage patient misreports while not compromising the fundamental abilities of MDS.

\noindent
\textbf{Interactive Experiment}\quad
For \textbf{RQ3}, mitigating misreports requires continuous dialogue to guide the patient in confirming or modifying the mentioned information.
We conducted interactive experiments utilizing a simulator-agent format.
Implementation details are provided in Appendix~\ref{appendix_interact}.

\begin{figure}[t]
  \centering
  \includegraphics[width=0.48\textwidth]{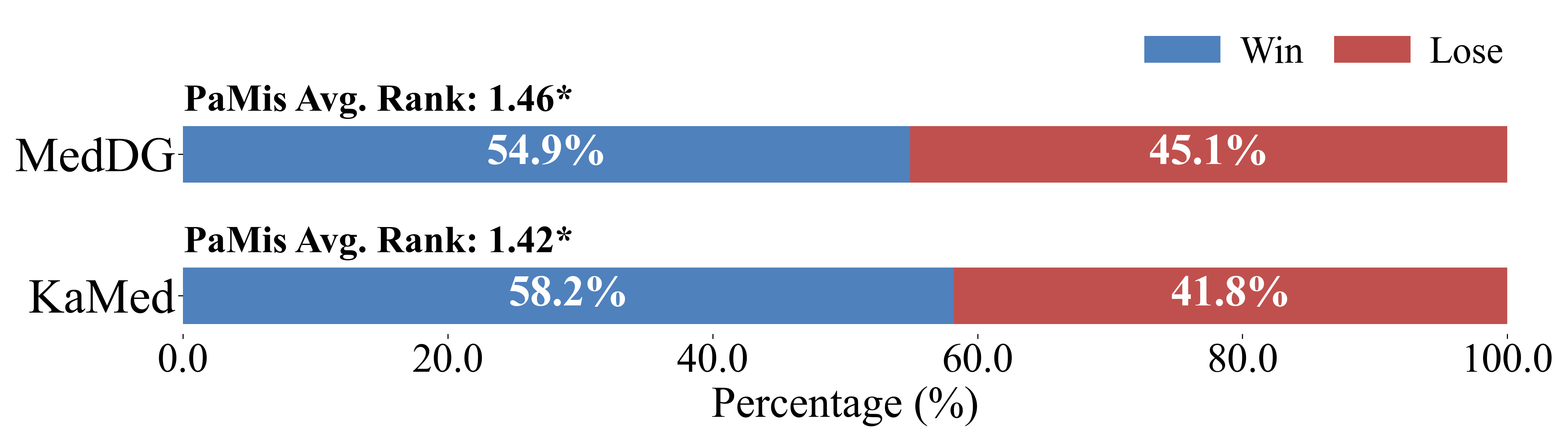}
    \caption{The human evaluation results of ~\modelname{} vs. GPT-4 (w/ Gold Know.) on two datasets.
 }
  \label{fig:human} 
\end{figure} 

\begin{table}[t]
    \tiny
        \centering
        \setlength\tabcolsep{2.5pt}
        \resizebox{0.49\textwidth}{!}{
        \begin{tabular}{l|ccccccc}
            \toprule[1.2pt]
            \textbf{Method}
            & \textbf{ B-1 }
            & \textbf{ B-4 }
            & \textbf{ R-1 }
            & \textbf{ R-2 }
            & \textbf{ D-1 }
            & \textbf{ D-2 }
            & \textbf{ E-F }
             \\
             \midrule[0.4pt]
            \multicolumn{8}{c}{\textbf{MedDG}} 
            \\\midrule[0.4pt]
            \textbf{ \modelname{} }
            &\textbf{44.28} &\textbf{24.88} &\textbf{28.12} &\textbf{13.80} & 1.23 &11.58 &\textbf{25.13}\\
            \midrule[0.4pt]
            \text{ w/o Detection } &37.06 &20.28 &27.05 &12.55 & \textbf{1.36} & \textbf{13.28} &24.02  \\
            \text{ w/o Mitigation } &42.17 &23.06 &28.05 &13.55 & 1.26 & 12.34 &24.61  \\
            \text{ w/o D\&M } &35.83 &19.57 &26.34 &11.83 & 1.29 & 12.48 &24.09  \\

             \midrule[0.4pt]
            \multicolumn{8}{c}{\textbf{KaMed}} 
            \\\midrule[0.4pt]

            \textbf{ \modelname{} }
            &\textbf{41.02} &\textbf{21.30} &\textbf{28.42} &12.01 &1.30 &11.39 &\textbf{28.18}\\
            \midrule[0.4pt]
            \text{ w/o Detection } &35.34 &18.36 &27.22 &11.95 & \textbf{1.34} & \textbf{12.70} &24.12  \\
            \text{ w/o Mitigation } &39.01 &19.74 &28.35 &\textbf{12.12} & 1.31 & 12.44 &24.89  \\
            \text{ w/o D\&M } &34.23 &17.99 &27.01 &11.37 & 1.33 & 12.36 &24.17  \\
            \bottomrule[1.2pt]
        \end{tabular}
        }
        \caption{Ablation study. ``D'' and ``M'' in ``w/o D\&M'' denote Detection and Mitigation modules respectively.}
        \label{tab:results_ablation}
    
    \end{table}

\definecolor{red0}{RGB}{192, 0, 0}
\definecolor{green0}{RGB}{0, 112, 192}

\begin{table*}[t]
\setlength{\belowcaptionskip}{-15pt} 
\small
    \centering
    \renewcommand\arraystretch{0.9}
    \resizebox{0.95\textwidth}{!}{
        \begin{tabular}{p{1.5cm}|p{12cm}}
        \toprule[1.5pt]
        \multirow{5}{*}{Context}
        & {\textbf{P}: After drinking a bit of liquor, I woke up in the middle of the night to use the restroom and noticed bleeding. The blood was bright red, and my stomach hurt a little.} \vspace{3pt}\\
        \multirow{5}{*}{}
        & {\textbf{D}: Hello, have you had any gastrointestinal issues in the past? \textbf{Do you have hemorrhoids?} How many times have you had bloody stools? Have you experienced abdominal pain, nausea, vomiting, fever, or any other discomfort?} \vspace{3pt}\\
        \multirow{5}{*}{}
        & {\textbf{P}: I have a slight \textbf{abdominal pain and a burning sensation}. I’ve \textbf{had discomfort from hemorrhoids before} but never treated it. I only had bloody stool once, and I haven’t had a bowel movement since. No other discomfort.} \vspace{3pt}\\
        \multirow{5}{*}{}
        & {\textbf{D}: Could you describe the location of your pain? If your gastrointestinal health has been good in the past, the most likely cause is an \textbf{acute hemorrhoid flare-up}, followed by \textbf{acute damage to the intestinal mucosa}.} \vspace{3pt}\\
        \multirow{5}{*}{}
        & {\textbf{P}: \textbf{The burning sensation is in the stomach area}, but there’s almost no pain.} \vspace{3pt}\\
        
        \midrule[0.6pt]
        Gold
        & {\textbf{D}: Alright, it's quite normal to experience a burning sensation in the stomach after drinking. How do you feel compared to yesterday?} \vspace{3pt}\\

        \midrule[0.6pt]
        Misreported
        & Heartburn: \color{red0}{Isolated} \vspace{3pt}\\

        \midrule[1.3pt]
        

        \midrule[0.4pt]
        GPT-4
        & {\textbf{D}: Your condition could be caused by {\color{green0}acute hemorrhoids or mild gastritis}. I suggest monitoring it for now, drinking plenty of water, and avoiding spicy foods. If the bleeding persists or the symptoms worsen, you may want to consider seeking medical attention.} \vspace{3pt}\\
        
        \midrule[0.4pt]
        \textbf{\modelname{}}
        & {\textbf{D}: Based on your description, the symptoms may be caused by {\color{green0}gastritis or hemorrhoids}. I recommend that you seek medical attention as soon as possible to determine the appropriate treatment plan. Have you experienced {\color{green0}acid reflux} recently?}\\
        
        \bottomrule[1.5pt]
    \end{tabular}}
    \caption{Case study on the misreported entities and the responses generated by LLMs.}
    \label{tab:case_study}
\end{table*}

The results of the interactive experiment are presented in Figure~\ref{fig:interact}.
It demonstrates that in both quantitative and qualitative analyses, the responses generated by models guided by \modelname{} exhibit the ability to mitigate misreporting.
This is reflected in an increase in the entropy of the entity graph, representing the completion of co-occurrence relationships, which leads to a more reasonable collection of symptoms.
Notably, after removing the knowledge related to misreported entities retrieved by PaMis, the results on MedDG perform worse than the baseline, suggesting that the naive retrieval method described in Section~\ref{sec:mitigation} is effective for mitigating misreporting. In contrast, the MR on KaMed slightly increased, possibly due to the longer average dialogue length in KaMed, which provides sufficient entities to be confirmed, thereby reducing the need for redundant retrieval.

Experimental results above indicate that \modelname{} has the potential to serve as an aid tool to alleviate the burden of doctor inquiries and mitigate subsequent risks of diagnostic mistakes.

\noindent
\textbf{Human evaluation} We selected GPT-4 (w/ Gold Know.)  as a baseline for comparison with the proposed method and conducted human evaluations using sample-wise comparisons.
We employed three doctors to assess the overall quality of the responses, with prior knowledge of the patient misreporting phenomenon.
As shown in Figure~\ref{fig:human}, ~\modelname{} significantly outperforms GPT-4 in ranking and achieves a higher sample win rate.


\subsection{Ablation Study}
\label{sec:ablation}
We conducted ablation experiments on the misreport-related modules to analyze their effects on response generation, as shown in Table~\ref{tab:results_ablation}.
It observed a decrease in E-F scores after removing the misreport detection module and employing a naive prediction method based on entity co-occurrence relationships (i.e., ``w/o Detection'').
Furthermore, after removing the detection module, the Distinct-1/2 scores of the generated results on both datasets increased. This may be because the generated content was not constrained by specific entities.
Removing the misreport mitigation module (i.e., ``w/o Mitigation'') resulted in fluctuations in BLEU and ROUGE scores.
An increase in R-2 can be observed on the KaMed dataset, which may be because the model no longer receives an excessive number of entities for constructing clarifying questions, resulting in outputs that more closely resemble the conversational style of doctors.
Further ablation results are provided in Appendix~\ref{appendix_ablation}.

\subsection{Case Study}
\label{sec:case}

As shown in Table~\ref{tab:case_study}, when the patient mentions ``{\color{red0}burning in the stomach}'', given that the patient's description may not be accurate, it could indicate either gastric mucosal damage or heartburn. Since there is no additional information supporting the heartburn symptom, it is necessary to ask the patient again about the presence of acid reflux to determine whether he is experiencing esophageal damage.
Existing models often ignore the potential inaccuracy, failing to confirm the patient's actual condition.
\modelname{}~retrieves related entities [Heartburn, Hemorrhoids, Gastritis] for response and probes the intermediate node ``{\color{green0}reflux}'', which could link heartburn to gastritis.

\section{Conclusion}
\label{conclusion}
This paper focuses on the differences between real patients and the typically assumed ideal users in the field of medical dialogue systems.
We defined the phenomenon that the content mentioned by the patient does not align with known health conditions as \textit{Patient Misreports} and we highlighted the importance of addressing this issue.
We then proposed~\modelname~to detect and mitigate patient misreports.
Experimental results indicate the high effectiveness of~\modelname~in response generation.
Interactive experiments further show its effectiveness in mitigating misreports.
We believe the proposed approach can serve as a diagnostic aid tool to alleviate the burden on doctors during consultations.

\section*{Limitations}
\label{limitation}
We first need to state that \modelname{} cannot independently work as a medical dialogue system, but is used to assist various generation models, including LLMs.
Secondly, the reliability of \modelname{} derives from analyzing co-occurrence relationships among entities within authentic corpora.
Hence, it is necessary to state that when the external knowledge base is fixed and not updated, \modelname{} cannot solve the problem of diseases and symptoms that are not included in the knowledge base, such as extreme cases absent.
Finally, given the critical importance of reliability in the medical field, this method should be used solely as an auxiliary tool to support doctors during consultations. To prevent potential harm to patients, it should not be employed as a direct diagnostic tool.

\bibliography{custom}

\clearpage

\clearpage
\newpage
\appendix

\section{Supplementary derivation of graph entropy}
\label{appendix_entropy}

\begin{figure}[t]
  \centering
  \includegraphics[width=0.49\textwidth]{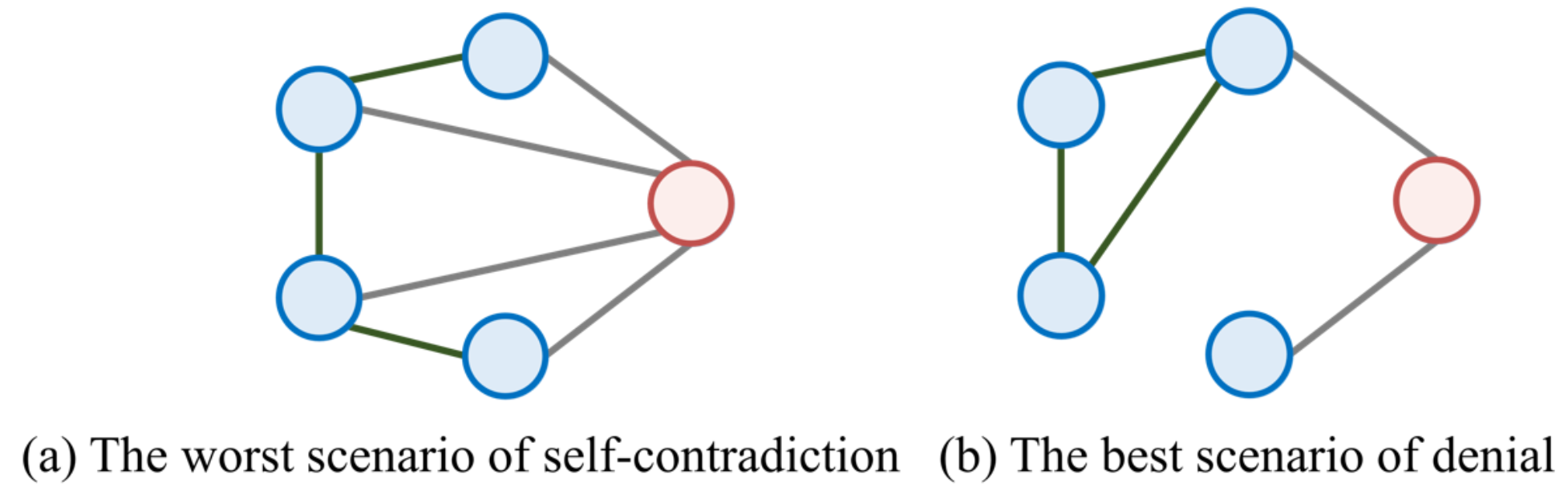}
    \caption{Different scenarios after losing a node (using the example of 4 remaining nodes).}
  \label{fig:graph_entropy} 
\end{figure}

If there are $n$ nodes remaining in the entity graph after a misreport, a self-contradiction involves at least $n-1$ edges connecting these nodes.
As illustrated in Figure~\ref{fig:graph_entropy}, in the worst-case scenario, the missing node was connected to every node in the original graph.
Therefore, the minimum graph entropy is given by:
\begin{equation}
    -\sum_{i=1}^{n} \frac{d_{i}-1}{2(n-1)} \log _{2} \frac{d_{i}-1}{2(n-1)}.
\end{equation}

In the case of a denial, the best-case scenario occurs when the missing node in the original graph has only two edges.
After the denial, the result is a connected graph with $n-1$ nodes and one isolated node, where the degree of the connected graph is $\operatorname{Vol}(G)-4 \geq 2(n-1)$.
Thus, the upper bound of entropy is:
\begin{equation}
    \begin{array}{r}
    - \left( \sum_{i=1}^{n-2} F(d_i) + F(d_{n-1} - 1) \right),\\
        F(d) =\frac{d}{\operatorname{Vol}(G) - 4} \log_{2} \frac{d}{\operatorname{Vol}(G) - 4}.
    \end{array}
\end{equation}

Subtracting them yields a new equation that is evident when $n=2$.
For $n \geq 2$, the left-hand side is a monotonically increasing function with respect to $n$, while the right-hand side remains constant, indicating that if the equation holds for $n=2$, it will also hold for any $n \geq 2$.
Consequently, the lower bound for contradictions surpasses the upper bound for denials:
\begin{equation}
\begin{array}{r}
    -\sum_{i=1}^{n} \frac{d_{i}-1}{2(n-1)} \log _{2} \frac{d_{i}-1}{2(n-1)} + \sum_{i=1}^{n-2} F(d_i)
    \\
    > - F(d_{n-1} - 1).
\end{array}
\end{equation}

\section{Baselines}
\label{appendix_baselines}
\textbf{LLMs}: (1)~\textbf{DISC-MedLLM}~\citep{bao2023discmedllmbridginggenerallarge}, a trustworthy medical LLM adapted for multi-turn dialogues.
(2)~\textbf{HuatuoGPT2}~\citep{chen2024huatuogptii}, an advanced medical LLM trained on Chinese medical corpus.

\textbf{Fine-tuned models}: (1)~\textbf{VRBot}~\citep{li2021semi}, a medical dialogue generation model based on patient entity tracking and doctor entity learning; and (2)~\textbf{DFMED}~\citep{xu-etal-2023-medical}, this framework performs the generation by fine-tuning a backbone model with dual-flow modeling.
(3)~\textbf{EMULATION}~\citep{xu-etal-2024-reasoning}, this framework relies on diagnostic reasoning analyses and aligns with clinician preferences through thought process modeling.

\section{Complementary Implementation Details}
\label{appendix_implement}

For comparisons with fine-tuned baselines on the response generation task, we follow the settings of previous works, allowing any doctor's response to serve as the target sentence.
We employ AdamW optimizer~\citep{loshchilov2017decoupled} with a weight decay of 0.02.
Reproducibility is ensured by fixing random seeds.
The proposed model is trained on an A100.
Training is conducted with a learning rate of 1e-4 for 10 epochs, with no adjustment of other hyperparameters apart from setting the maximum generation length to 160 tokens.

\begin{table*}[t]
    \centering
    \begin{tabular}{c|p{0.8\textwidth}}
        \toprule[1.5pt]
        
        \multirow{5}{*}{\makecell[l]{Doctor}}
        & You are a doctor conducting a conversational consultation with a patient. \\
        & \textbf{1.} Take the information from the [Dialogue History] into account, which may include incomplete, inaccurate, or inconsistent details in the patient's statement. \\
        & \textbf{2.} Reference the 'Potentially Inaccurate Entity' and 'Related Entity' in [Medical Knowledge] to provide accurate medical advice and help resolve uncoordinated issues. \\
        & \textbf{3.} Respond in a way that is concise, approachable, and compassionate. Ask follow-up questions to gather more details and may also correct errors. \\
        & \textbf{4.} Keep the conversation natural, focusing on one or two key points at a time to ensure the patient feels supported and informed. \\
        & \textbf{5.} The response should be bite-sized and not give too much information at once, which is similar to what the doctor did in dialogue history. \\

        \midrule[1.2pt]

        \multirow{4}{*}{\makecell[l]{Patient}}
        & You are a patient engaging in a conversational consultation with a doctor. \\
        & \textbf{1.} Consider the [Dialogue History] and [Patient's Symptoms] for context to continue the conversation naturally. \\
        & \textbf{2.} Respond briefly, using one to two short sentences. Describe your symptoms in a vague and colloquial manner, avoiding professional terminology. \\
        & \textbf{3.} Express emotions like worry, anxiety, or hesitation as appropriate to reflect your life experiences. \\
        & \textbf{4.} Keep your responses natural and reflective of how a patient might realistically communicate their concerns. \\
        
        \bottomrule[1.5pt]
    \end{tabular}
    \caption{The prompts for simulated patients and doctors.}
    \label{tab:prompts}
\end{table*}

\section{Interact Settings}
\label{appendix_interact}
Addressing concerns about the quality of dialogue simulators~\citep{wang2024indepth}, we adopted the method proposed by \citet{chen2023llmempowered} to employ LLMs as simulated patients.
The interactive experiment was initiated by selecting instances where misreports were detected within the dialogue.
We then input the dialogue context into LLMs and requested that they respond as patients, guided by the generated doctor's inquiries.
We employed the graph entropy method to assess whether misreports within the dialogue were mitigated.
Notably, although the ChatGPT series models are not specialized in the medical domain, they surpass most models in medical-related evaluations\footnote{\url{paperswithcode.com/sota/question-answering-on-medqa-usmle}}. 
We build on the prompt settings provided by~\citet{chen2023llmempowered} to simulate patient-doctor interactions, as illustrated in Table~\ref{tab:prompts}.
Simulated patients are utilized to assess whether doctor responses can help patients clarify their actual health conditions, thereby mitigating misreports.
The prompts above were constructed with the participation of medical professionals and demonstrated their effectiveness as simulations of patients and doctors.

\begin{table*}[ht]
\tiny
    \centering
    \resizebox{0.95\textwidth}{!}{
    \begin{tabular}{l|ccc|ccccc}
        \toprule[1.2pt]
        & \textbf{ P }
        & \textbf{ R }
        & \multicolumn{1}{c|}{\textbf{ F1 }}
        & \textbf{ F1\(_A\) }
        & \textbf{ F1\(_D\) }
        & \textbf{ F1\(_E\) }
        & \textbf{ F1\(_M\) }
        & \textbf{ F1\(_S\) }
        \\
        \midrule[0.8pt]
        \bm{$Is$} &34.48&44.44&\multicolumn{1}{c|}{38.83}&100.0&50.00&0.0&35.90&30.49\\
        \bm{$Is$}  \scalebox{0.75}{w/o Detection}&30.32&41.05&\multicolumn{1}{c|}{34.88}&100.0&25.00&50.00&31.97&29.64\\
        \midrule[0.4pt]
        \bm{$De$}&43.78&47.07&\multicolumn{1}{c|}{45.37}&57.32&61.11&33.33&40.41&36.24\\
        \bm{$De$} \scalebox{0.75}{w/o Detection}&41.76&48.66&\multicolumn{1}{c|}{44.95}&54.37&70.32&33.33&32.31&40.85\\
        \midrule[0.4pt]
        \bm{$Co$}&30.21&43.86&\multicolumn{1}{c|}{35.78}&50.00&42.86&100.0&70.59&29.73\\
        \bm{$Co$} \scalebox{0.75}{w/o Detection}&16.99&37.27&\multicolumn{1}{c|}{23.35}&45.45&68.18&0.0&50.00&16.00\\
        \bottomrule[1.2pt]
    \end{tabular}}
    \caption{Ablation results for entity prediction on samples containing different types of misreports. A-S denote attribute, disease, examination, medicine, and symptom.}
    \label{tab:results_ablation_hal_type}
\end{table*}
\section{Supplementary Ablation Study}
\label{appendix_ablation}

As illustrated in Table~\ref{tab:results_ablation_hal_type},~\modelname~outperforms the overall average on samples containing misreports, and the detection method is effective for all types of misreports, resulting in an improved F1 score when perform entity prediction task. The performance only declines on disease entities, possibly due to the tendency to select other types of entities to fill co-occurrence relationships rather than making direct diagnoses.

\end{document}